# Optimization meets Big Data: A survey

*Ricardo Di Pasquale*
Facultad de Ingeniería y Ciencias Agrarias
UCA (Pontifica Universidad Católica Argentina)
Buenos Aires, Argentina
rdipasquale@uca.edu.ar

*Javier Marenco*
Instituto de Ciencias
UNGS (Universidad Nacional de Gral. Sarmiento)
Buenos Aires, Argentina
jmarenco@ungs.edu.ar

*Abstract —* This paper reviews recent advances in big data optimization, providing the state-of-art of this emerging field. The main focus in this review are optimization techniques being applied in big data analysis environments. Integer linear programming, coordinate descent methods, alternating direction method of multipliers, simulation optimization and metaheuristics like evolutionary and genetic algorithms, particle swarm optimization, differential evolution, fireworks, bat, firefly and cuckoo search algorithms implementations are reviewed and discussed. The relation between big data optimization and software engineering topics like information work-flow styles, software architectures, and software framework is discussed. Comparative analysis in platforms being used in big data optimization environments are highlighted in order to bring a state-or-art of possible architectures and topologies.

*Keywords – Big Data; Optimization; Metaheuristics*

## I. INTRODUCTION

Optimization is a field at the intersection of mathematics, computer science, and operations research, which studies methods to find the best solution(s) according to a criterion usually given by an objective function, among a set of alternatives bound by a set of constraints.

The big data concept is related to the theory and technologies that allow the processing of big volumes of data, which, for size or complexity reasons, cannot be processed with traditional tools. The word "big" as a prefix means more than an evolution, and it's a paradigm shift instead: Classical data analytics cannot deal with the big data 5Vs [1]-[2] (volume, velocity, variety, veracity, and value).

Data science is considered an evolutionary extension of statistics [3] with the added capacity of dealing with massive amounts of data. It is considered a fusion between computer science and statistics [1].

Big data analytics (BDA) is an evolution of data analytics (and data mining) techniques and algorithms [4]. A comprehensive definition of big data analytics is given in [5] [6]: *"Big data analytics is the area of research focused on collecting, examining, and processing large multi-modal and multi-source data sets in order to discover patterns, correlations and extract information from data"*.

Although data mining and optimization are different fields of study, there are many points in common, e.g., a classification problem can be considered an optimization problem where the goal is to maximize the classification accuracy and minimize the complexity under certain constraints [7]. In this vein, machine learning (ML) and optimization can be considered as two faces of the same coin [2]. Knowledge discovery in databases (KDD) problems can be treated as optimization problems e.g. [8]-[11].

What happens when optimization meets big data? Is there a new field of study at this intersection? If optimization algorithms could be applied to big data with no additional effort or study, then there would be no need for a new field. Also, a new field of study would not be necessary if any of the following elements do not make any effect on optimization results: new data types, new data ingest patterns, streaming, data complexity or data size. How optimization interacts with big data is discussed in this paper.

Several works in the literature like [12]-[17] contributed to consider Big Data Optimization (BDO) as a new field. BDO arises where new optimization algorithms (or scalable versions of classical algorithms) and techniques must be developed to be applied in combination with Big Data Analytics (BDA). This way, BDO arises from a *cross-fertilization* [18] between statistics, optimization, and applied mathematics.

BDA aids BDO in dealing with data analysis. BDO employs all the tools used in classical optimization, but the most distinctive feature of BDO is the development of distributed and scalable implementations of classical optimization algorithms, as much as the development of new methods and algorithms to deal with big data issues [15].

The remainder of the paper is organized as follows. Section II reviews previous work on exact methods including integer linear programming, classical optimization solvers, and coordinate descent methods introducing hadoop map reduce as one of the main big data platforms. Section III focuses in the use of metaheuristics, simulation, and iterative methods in BDO. Section IV discusses the role of software engineering in BDO, as well as state-of-the-art BDO information work-flow styles, software architectures and frameworks. Section V reviews the main platforms, while the conclusions and future trends are drawn in Section VI.

## II. BIG DATA OPTIMIZATION IN EXACT METHODS

The first open source framework for the implementation of distributed file systems and big data processing via the *map reduce* (MR) framework [19] was Apache Hadoop [20].

The work in [21] integrating Integer Linear Programming (ILP) denotes the difficulty of efficiently solving ILP in this context, and literally asks *"can we find an easy way to handle the massive computations without involving sophisticated mathematical logic?"* The rest of [21] advocates to answer this question by stating that the Hadoop framework is suitable to run these kind of solutions. The solution in [21] is based on applying a dual decomposition method to separate variables of an ILP problem in a smaller subset of ILP dual subproblems (i.e., of a smaller dimension), combined with the notion of iterative updates in Lagrangian multipliers can fit into Hadoop's MR model. The authors implemented an air traffic flow optimization problem of the USA National Airspace System (NAS) to illustrate their statements. IBM Ilog CPLEX Java API was used to integrate to a Hadoop MR cluster. Distributed processing was handled by Hadoop, and parallel local processing (ILP subproblems) was handled by the CPLEX solver. Numeric results are very interesting, and denote scalability, which is an appreciated distributed processing property. Although it's an interesting approach, well implemented and documented solution, there are a few observations to have in mind: (a) Hadoop MR is well suited to process big amounts of data in a streaming way, which excludes the iterative nature of process. If every decomposed subproblem could not be identified in only one iteration, the overhead of the framework could be traumatic. And (b), the classical Hadoop MR approach is to "send" algorithms where data is allocated. If the subproblems cannot be fulfilled by data allocated in their node, then the network impact may also be traumatic.

Another approach to implement distributed optimization is [22], where the objective is to introduce a framework for implementing distributed optimization with arbitrary local solvers. This work extends the CoCoA framework [23] with a new version called CoCoA+. In [22] is stated that traditional optimization solvers –the ones working in only one node– have been developed and improved for a long time by the software industry, and, in consequence, nowadays perform better than the new distributed solvers. Opposite to [21], [22] considers Hadoop MR a too slow hard disk-based work-flow system. CoCoA+ is implemented in C++/MPI. Numeric results show that CoCoA+ converges faster than other algorithms like DiSCO [24], but tends to be slower after some iterations. Performance analysis comparing local solvers are also shown in [22]. Some observations follow: (a) data distribution needs some prerequisites in order to run under MPI. Fault tolerance needs to be implemented manually in C++/MPI. And (b), the observation about data availability in each node in order to be processed by local solvers made to [21] is valid in [22].

Coordinate descent methods (CDM) are iterative optimization algorithms that deal with an optimization problem by solving a sequence of lower dimensional optimization problems. It's considered one of the more successful algorithms in BDA and BDO environments [4]. Essentially, the algorithm tries to modify only one coordinate of the variables vector for each iteration. There are parallel and distributed implementations of these algorithms. In the work proposed in [4], LASSO algorithm was implemented with CDM, showing very interesting numeric results, and a near linear speedup. The experiments were executed on 24 cores. Regrettably, the work does not include multinode distributed executions of the algorithm.

## III. BIG DATA OPTIMIZATION IN HEURISTIC METHODS

After Hadoop MR has become the main big data processing platform, the main problems detected were: (a) Lack of iterative nature: Hadoop MR does not admit an iterative style of programming, and implementing such an algorithm implies overhead that limites development. There were many documented efforts to avoid this non-iterative nature like [25] and [26]. And (b), limited power of expression of the map reduce style. There were efforts to make developers "think in map reduce" like [27], where graph algorithms are implemented (and extensively explained) in a pure map reduce manner. However, in general state cannot be simply shared between mappers, and it is not simple to convert algorithms to this format.

When Apache Spark was introduced, it rapidly turned into the evolution of the Hadoop MR framework. It is built upon the implementation of Resilient Distributed Data (RDD) as introduced in [28], to build in-memory clusters. This kind of structures could take advantage of all the benefits of Hadoop and could also bring iterative nature to the process and expand framework expressive power. Most of this kind of optimization needs a support for iterative implementations. The emerging of Spark as a standard for big data processing was great news for BDO.

### A. Alternating direction method of multipliers (ADMM)

Originally developed in 1974 [29], this method was introduced to solve convex optimization problems by dividing them into smaller pieces. Although ADMM could be considered an exact method, its nature allows implementations (modifications) well suited for delivering approximated results. This is an appreciated feature in big data environments.

The implementation of [15] aims to apply ADMM-based algorithms to the optimization of communications smart grids, particularly power flow systems with security constraints. It introduces the ADMM technique in two blocks, and an extension to $n$ blocks. The authors affirm that ADMM has a parallel nature that can be implemented with Spark, excluding Hadoop MR because its problems with iterative procedures. The authors referenced a previous publication [20] where they proposed a distributed parallel approach for a big data scale optimal power flow with security constraints. The goal of [15] is to optimize economics in the power flow of electric energy dispatch introducing the evaluation of security constraints or risk analysis scenarios. This implementation takes the classical economic dispatch problem and introduces some previously elaborated contingency scenarios as constraints. The implementation is based in ADMM distributed process in a way that each node can solve a different subproblem in parallel. The numerical results showed that after a few iterations, the output of the proposed algorithm were near to optimal solutions, which means that the solution is able to approximate good solutions rapidly. They used a standard benchmark (called IEEE 57 bus), which shows a speedup factor within the interval (1.4, 2.4), but it is supposed to raise the speedup factor within the interval (4.4, 4.8) by improving communications.

### B. Metaheuristic algorithms

*1) Evolutionary algorithms (EA):* EA, including genetic algorithms (GA) is a field that naturally can make use of distributed and parallel computing. The work described in [1] focuses on GA and swarm intelligence. The authors vision with respect to the simultaneous application of BDA and EA should be considered dual: apply EA on big data, or apply big data techniques in order to improve EA.

The authors of [1] emphasize the difference in methods used to generate new solutions in metaheuristics, recognizing two main categories: instance based search (IBS), and model based search (MBS). In IBS new candidate solutions are generated using only the current solution. Simulated annealing and iterated local search belongs to IBS category. In MBS the candidate solutions are generated using an (explicit or implicit) model that employs information of previous candidate solutions, making the search focus in high quality solution regions. Ant colony optimization and estimation of distribution algorithms belong to the MBS category. Finally, [1] establishes key challenges of EA in order to sole BDA problems:

- BDA requires rapid data mining on big volumes of data. It reinforces the idea of characterizing data mining problems as optimization problems.

- The complexity of data is also relevant, given that high-dimensional data is not the same that a big volume of data. Performance in optimization problems uses to decrease exponentially according to the number of variables or goals, as far as the search spaces grows (also exponentially).

- Dynamic problem management: data nature in real world is dynamic, changes rapidly in short periods of time. Sometimes this kind of problems can be considered non-stationary environment [30] or uncertain environment [31] problems, in both cases EA were applied successfully [32] [33].

- Multi-objective optimization: EA are particularly good on this kind of problems, allowing to find a Pareto optimal set in only one run [34].

A big data perspective on EA is given in [35], where it is stated that in order to explore a big data store (and also high-dimensional data) some kind of data analysis is needed to guide genetic operations. The use of informed genetic operators (IGOs) is illustrated in order to achieve a good performance in big data EA (or GA). IGOs usually use a meta-model (or reduced model) that is built on top of data analysis. In [35] the use of IGOs is aligned with the idea of expanding the search to non-explored regions. These kind of regions are discovered by allowing not well-suitable individuals to develop. It implies that the flow of GA (or EA) must not kill those kinds of individuals prematurely. In order to achieve this goal, populations are divided in tiers of sub populations built with the idea of being dynamic niches classified by fitness ranks where promising sub-optimal solutions could survive. Its implementation [35] is very similar to the multiple-deme GA approach in [36], in particular, the idea of population islands and migration process. The algorithm avoids local convergence by identifying it in the algorithm work-flow and replacing redundant individuals (minor fitness criteria) by more promising candidate solutions from unexplored areas of the solution space. The IGO operation in [35] is a mutation.

*2) Parallel Particle Swarm Optimization:* The PSO approach based on Hadoop MR was used in [37] to optimize a back propagation (BP) neural network. The goal of this work is to improve the precision and performance of the classification. The implementation, based in Hadoop MR, makes use of a classical PSO algorithm with a parallel design. Authors highlight how optimization techniques (algorithms and metaheuristics like PSO and GA) have been used successfully to adjust neural networks weights. Besides, this work values research works focused in developing parallel or distributed designs for classical algorithms as, e.g., [38] where a MPI implementation for a BP neural network algorithm in a supercomputer was developed. Some of these works rely on GPU technology, which require a detailed knowledge of the hardware where the solution runs, highlighting Hadoop MR as a hardware independent solution. Some works in Hadoop MR were referenced in [37], like [39] where a MR model was used to design a density-based clustering algorithm with good experimental results.

One of the most important problems in migrating BP algorithms to big data platforms referenced in [37] is how inadequate traditional attribute reduction (fundamental in sets theory) perform, and in order to deal with it, authors reference [40], where hierarchical attribute reduction algorithms for big data using MR were proposed.

### C. Simulation optimization (SO)

SO studies how to find optimal solutions to systems represented in computer simulation models. It has some distinctive characteristics [42]: objective values can only be estimated with certain noise level and an important computational cost, given that each objective function evaluation requires a simulation run.

SO has been used in several environments, given the complex nature of models of real stochastic systems that cannot be processed with other techniques.

The authors emphasize the advantages of cloud computing related to SO, remarking possibilities brought by parallelism and distributed processing. The authors state two types of challenges presented by big data adoption in SO: large data, and the quantity of existent data sources and the different styles of information processing. The authors suggest to adopt "multi-objective optimization with ordinal transformation and optimal sampling" MO$^2$TOS [43], a framework for SO running as far as data processing.

### IV. Big data optimization and software engineering

As stated in [18], BDO emphasizes the idea of a cross-fertilization between statistics, applied mathematics, and optimization, but it seems to be insufficient: BDA and software engineering (SE) need to be heavily considered when implementing algorithms in this field.

As seen on [13], [14], and [17], researchers had to build software frameworks to deal with BDO. It seems to be evident that the complexity of big data not only affects the "data" dimension of the BDO solutions. Usually, optimization researchers do not need do to deal with software frameworks, but BDO researchers cannot ignore this need. Something similar happens with the "process" dimension: optimization researchers just needed the data in order to use it as an input for the optimization model, but BDO interacts with processes and data in a different way, a much more integrated way. Hence, the big data paradigm shift in optimization implies to deal not only with big databases or high-dimensional data, but also with work-flows and software frameworks.

As seen in [2], it is a good practice to tackle a complex system by dividing it in less complex parts, in order to optimize them sequentially. With BDO, optimization problems need to be tackled in an integrated way.

In particular, software architecture turns out to be a relevant issue in this cross-fertilization schema, in order to develop better BDO solutions and to be able to apply optimization to new types of data like, e.g., stream data [13] [44].

To the best of our knowledge, there are no specific publications on BDO SE, but there are several interesting works in big data SE like [45]-[48]. In BDA SE, the conference paper [47] appeared at the 2017 IEEE CCWC in BIGDSE [48].

The "process" dimension, usually studied in SE, adds value to BDO by specifying the information work-flow style to be used in a particular solution. There are no studies of information work-flow styles in BDO, but information extracted from the study of the bibliography suggests the following classification.

## A. A classification of BDO Information work-flow styles

### 1) Classical cascade work-flow

The simplest work-flow to implement. It starts with the BDA phase, where the data sources are processed in order to build the optimization model input. Then, the optimization model (e.g., LP) produces results or solutions. The particle swarm algorithm implementation shown in [37] implements this type of work-flow, as well as the airline route profitability analysis and optimization in [41]. The diagram in "Fig.1" represents the classical cascade work-flow.

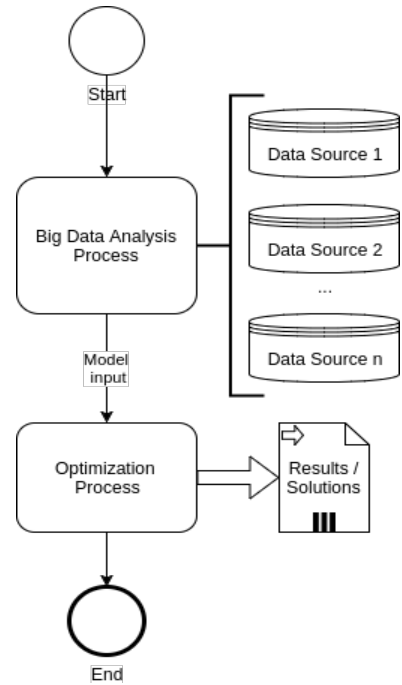

Fig. 1. Classical cascade work-flow

### 2) Iterative cascade work-flow

This style of work-flow is similar to the classical cascade work-flow. The iteration may happen when (a) a refinement is required in the data process, or (b) streams processing may refine the data process because of its real time nature. More refined results are output from the optimization process at each iteration. The simulation methods for optimization [42] employ this style of work-flow. The diagram in "Fig. 2" represents this style.

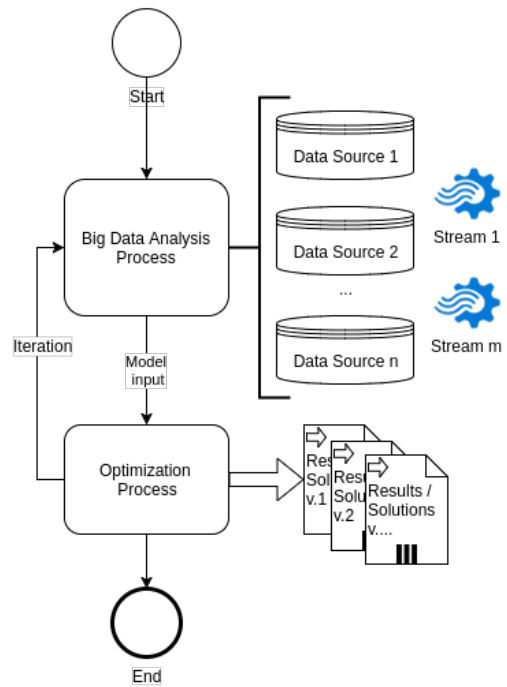

Fig. 2. Iterative cascade work-flow

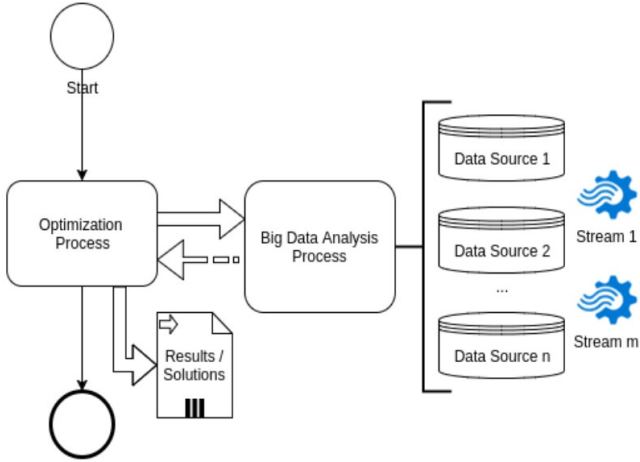

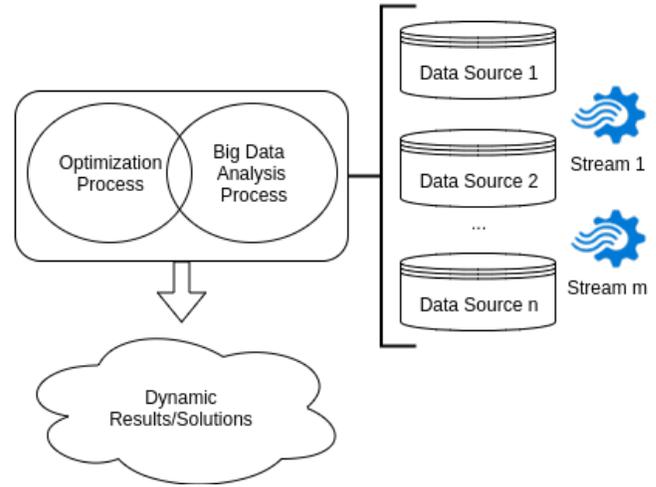

Fig. 3. Feedback model work-flow

Fig. 4. Integrated model

### 3) Feedback model work-flow

This style of work-flow is designed under the premise of the optimization model being the main process of the solution, and implementing a producer-consumer pattern style, where BDO is the producer and the optimization model is the consumer. The Fireworks Algorithm Framework in [14] works under the premise of an information work-flow similar to the feedback model work-flow, as well as the work in [49] and [50]. The diagram in "Fig. 3" illustrates the feedback model work-flow.

### 4) Integrated model

This model is not a work-flow at all, just because both processes are conceived fully integrated to each other. In this model, the feedback between optimization and BDA is continuous, and the results are conceived as dynamic.

When streaming in real time is the main data source, and the model must solve in near real time, this work-flow model should be employed, also when the "best solution possible" approach is needed in a real time context. The diagram in "Fig. 4" illustrates this model.

### B. Software architecture (SA)

Developments in SA like Lambda Architectures [51] allow BDO to implement solutions with sophisticated BDA on-line and in a near real time manner. In this kind of architectures, the "best result right now" approach is preferred. Usually the dynamic nature of the BDO problems would make obsolete results in short periods of time. Although there are many critics to Lambda Architecture like [52], other software frameworks to provide similar solutions are still developing, like Apache Samza and Apache Beam. Researches in SA are worried about how cloud computing can serve as a platform for enterprise integrated big data solutions. An interesting road map is documented in [48].

### C. Software frameworks (SF)

SF are usually studied in SA, as part of SE. The value added by software frameworks to BDO allows optimization researchers to focus in "optimization" itself, dealing with data and process distribution and parallelization as well as domain model design. SA should be studied as high level conceptual frameworks.

One of the main ideas of SA is the reuse and extension of software and concepts. The authors of [17] extend jMetal, an existent multi-objective optimization framework implemented in Java [53]. This extension enables the possibility of running distributed metaheuristics on Spark. It also manages streaming Spark data sources. In order to validate the new extension, authors have implemented a bi-objective TSP [54] with real traffic flow data. They choose a GA called NSGA-II (Non-dominated Sorting Genetic Algorithm-II), extending provided jMetal functionality in order to dynamically adjust several execution parameters.

In Section III we mentioned that [14] presents a FWA framework. This framework groups several implementations of FWA, and sets some domain entities suitable for extending the framework by adding another FWA. There is not enough documentation in [14] to evaluate this SF from a SA perspective. Judging its utility by studying [14], it is a reasonably good toolkit for FWA.

Something similar occurs with [13]. Also introduced in Section III, it presents an automated differential evolution SF (ADEF). Its main objective is to encapsulate DE algorithms and variants (e.g., multi-objective) logic, but it automatically configures the best set of operators to use, as no single DE operator is considered the best for solving all types of optimization problems. There is not enough information in [13] to evaluate the SF from a SE perspective.

Mentioned in Section II, CoCoA+ was developed in [22] as an extension of the CoCoA framework in C++/MPI with the objective of managing the distributed processing of local arbitrary solvers running on a cluster. Observations from the SE perspective were presented in Section II. Too much work is needed to efficiently distribute data, manage process, and synchronize state in MPI with arbitrary tools running locally all over the cluster.

Mentioned in Section II, MO²TOS was presented in [43] and used in [42] to run SO. This SF has two main objectives: (1) support SO modeling and (2) support BDA data processing activities. The SF consists of two methodologies: ordinal transformation (OT low fidelity) and optimal sampling (OS high fidelity). No details about its implementation are provided in the bibliography, so it is impossible to find out its internal architectural characteristics.

## V. Big Data Optimization Platforms

The focus in this section is to review useful BDA platforms in the BDO context.

The Hadoop file system (HDFS) is one of the most important standard on distributed file systems for big data purposes. A great early and extensive work about HDFS from Yahoo researchers is given in [55].

MPI is a portable library designed to provide synchronization and communication functionality to distributed processes. It specification was standardized in 1993. It is well suited for supercomputer architectures and high performance network clusters. Nowadays its main application is to process heavy CPU intensive solutions [56]. When used with commodity hardware, it can be combined with Beowulf clusters [57], which is a cluster architecture specification compatible with low cost clusters and open source software.

As mentioned in Section II, Hadoop MR was introduced as a simplified and highly scalable data processing platform [19] well suited for BDA processing. There are many higher abstraction level tools that are based on MR. Its main advantages on the big data context over previous technologies like MPI are discussed in [27]:

- The way the problem is divided in smaller parts in order to be executed in a distributed platform: Hadoop MR abstracts these processes.

- The way tasks are assigned to workers in a cluster: In MPI it must be explicitly defined by developers. Hadoop MR leverages the assignment of map or reduce task in agreement with historical performance.

- In MPI, data must be allocated carefully in order to let the workers have all they need to process. Hadoop MR employs a different paradigm: algorithms travel through the network to where the data is allocated, not the other way around.

- The way synchronization between nodes is coordinated: MPI developers usually specify a master program. In Hadoop MR there are a job and a task coordinator implemented in the core of the platform.

- The way high availability (HA) and fault tolerance (FT) are achieved: MPI HA and FT has to be programmed by developers, or to be provided by cluster. Hadoop MR is designed to provide HA and FT.

MR disadvantages include the lack of iterative nature and a limited power of expression. Spark emerged as a solution for both MR disadvantages. Based on RDD structures, Spark enables in-memory cluster processing [28]. It also includes several actions and transformations that facilitate BDA processing. It also includes specific ML, streaming, and graph processing libraries.

Several works have compared distributed and big data platforms. Specifically for the BDA context, [6] is a comparative performance analysis for Spark on Hadoop against MPI/OpenMP on Beowulf clusters. The selected algorithms for performance benchmarking were KNN (K-Nearest Neighbors) and Pegasos SVM (Support Vector Machines). Data used for benchmark were 7Gb of HIGGS data [58] with 11 million records and 28 dimensions. Ref [6] highlights the difference in performance between Spark and Hadoop, placing Spark closer to the high performance computing (HPC) paradigm. In CPU terms, numeric results showed a superior performance in Beowulf, but, in a qualitative analysis, Spark advantages on FT, replication management, hot nodes swapping and simplicity. Several observations about the results shown in [6] must be considered: (a) Spark's own KNN and SVM algorithms were not used. MLLib provides such implementations, and it should overperform a simple MR implementation. (b) An advantage on Spark/Hadoop to process large data is shown. Big data platforms were designed to process many terabytes of data, even petabytes. The amount of 7 Gb should not be considered enough volume of data in order to compare big data platforms with HPC platforms, as this may bias the results.

Near real time BDA is an emerging field related to streaming. In this kind of solutions, analysis should not be based on batch patterns. As stated in [2] and [44], there are many cases where BDO must be applied directly on streaming data. A comparative analysis on streaming BDA performance is made in [44], where Spark Streaming capacities were compared with Apache AsterixDB use. AsterixDB is a scalable, open source big data management system that still has not reached version 1 release, but has many of BDA concerning features to be considered. In [44], the Spark/Cassandra was compared with AsterixDB in a social network data streaming solution. The algorithms used in [44] were "word count" and a sentiment analysis algorithm. Results in [44] show that AsterixDB reached a better throughput and lower latency. AsterixDB suffered some performance issues when data batch segments size was increased. The author remarked that the work was not accomplished by Spark or AsterixDB experts, and neither implemented in a functional language like Scala. Several discussion issues were documented after publication, e.g., Spark/Cassandra version stores streaming data before processing, which generates overhead, where a simple workaround or design improvement would increment Spark/Cassandra throughput.

## VI. Conclusions and Future Trends

BDO has emerged as a new field of study in last years. BDO emerges from a cross-fertilization environment between optimization, applied mathematics, statistics, BDA, and SE. Not always classical HPC platforms are suitable for running BDO software, but specific BDA platforms like Spark on Hadoop have demonstrated to be more accurate for this task. Several optimization algorithms have been successfully rewritten in order to be applied to BDO environments. Other algorithms have been developed specifically to match BDO requirements.

A new classification on information work-flow styles for BDO is proposed in this work, inspired by the review of the bibliography and previous studies.

Future trends include new comparative studies and benchmarks in specific BDO problems with accurate amounts of data; new applications in classical optimization problems as well as new industrial applications; studies on the implications of real time streaming BDO and the paradigm shift of continuous real time BDO.

REFERENCES

[1]  S. Cheng, B. Liu, Y. Shi, Y. Jin, and B. Li, "Evolutionary computation and big data: key challenges and future directions", in: Y. Tan, Y. Shi (eds) Data Mining and Big Data. DMBD 2016. Lecture Notes in Computer Science, vol 9714. Springer, Cham, pp. 3–14, 2016. DOI: 10.1007/978-3-319-40973-3_1.

[2]  A. Lodi, "Big data and mixed-integer (nonlinear) programming" - Data Science for Whole Energy Systems, Alan Turing Institute scoping workshop, 28-29 Jan 2016, Edinburgh.

[3]  D. Cielen, A.D.B. Meysmann, M. Ali, "Introducing data science. Big data, machine learning, and more, using python tools", Manning Publications Co., 2016, ISBN 9781633430037.

[4]  C.Tsai, C. Lai, H. Chao, and A. Vasilakos, "Big data analytics: a survey", Journal of Big Data (2015), a Springer Open Journal 2:21 DOI 10.1186/s40537-015-0030-3.

[5]  Y. Zhai, Y. Ong, and I. Tsang, "The emerging big dimensionality", IEEE Computational Intelligence Magazine, Vol. 9, Nr. 3, pp. 14–26, 2014.

[6]  J.L. Reyes-Ortiz, L. Oneto, and D. Anguita, "Big data analytics in the cloud: Spark on Hadoop vs MPI/OpenMP on Beowulf", Procedia Computer Science Volume 53, 2015, pp. 121–130, 2015 INNS Conference on Big Data.

[7]  W. Art Chaovalitwongse, C.A. Chou, Z. Liang, S, Wang, "Applied optimization and data mining", Ann Oper Res  249, pp. 1–3, DOI 10.1007/s10479-017-2402-x, 2017.

[8]  S. Ezhil, C. Vijayalakshmi, "An implementation of integer programming techniques in clustering algorithm", Indian Journal of Computer Science and Engineering (IJCSE), Vol. 3, No. 1 Feb-Mar 2012, pp. 173-179, ISSN: 0976-5166.

[9]  G. Pavanelli, M.T.A. Steiner, A. Goes, A.M. Pavanelli, D.M. Bertholdi Costa, "Extraction of classification rules in databases through metaheuristic procedures based on GRASP", Advanced Materials Research, VL – 945-949, pp. 3369-3375,  Trans Tech Publications, 2014.

[10] V. Ramasamy, "Prioritization of association rules using multidimensional genetic algorithm" in International Journal of Applied Engineering Research, Vol. 10, Nr. 55, pp. 2288-2291, January 2015.

[11] W. Wu, L. Liu, B. Xu, "Application research on data mining algorithm in intrusion detection system", Chemical Engineering Transactions, Vol. 51, 2016, ISBN 978-88-95608-43-3, ISSN 2283-92162016.

[12] A. Emrouznejad (Ed.), "Big data optimization: recent developments and challenges", Springer, ISBN 978-3-319-30263-8, DOI 10.1007/978-3-319-30265-2, 2016.

[13] S. Elsayed and R. Sarker, "Differential evolution framework for big data optimization", Springer, Memetic Computing, DOI: 10.1007/s12293-015-0174-x, 2016.

[14] M. El Majdouli, I. Rbouh, S. Bougrine, B. El Benani, and A. El Imrani, "Fireworks algorithm framework for big data optimization" - Springer, Memetic Computing 8(4): pp. 333-347, DOI: 10.1007/s12293-016-0201-6, 2016.

[15] L. Liu and Z. Han, "Multi-block ADMM for big data optimization in smart grid", Systems and Control (cs.SY), DOI: 10.1109/ICCNC.2015.7069405, 2015.

[16] P. Richtarik and M. Takac, "Parallel coordinate descent methods for big data optimization", Mathematical Programming, 156 (1). Springer. pp. 433–484, 2012.

[17] C. Barba-González et al., "A big data optimization framework based on jMetal and Spark" ("Un Framework para Big Data Optimization Basado en jMetal y Spark"), The 2nd International Workshop on Machine learning, Optimization & big Data, MOD 2016, SIAF Learning Village Tuscany – Volterra (Pisa), Italy, August 26-29, 2016.

[18] J. Fan, F. Han and H. Liu, "Challenges of big data analysis", Natl Sci Rev, 1 (2), pp. 293-314, 2014

[19] J. Dean and S. Ghemawat, "MapReduce: simplified data processing on large clusters", OSDI'04 Proceedings of the 6th conference on Symposium on Opearting Systems Design & Implementation, Vol. 6, pp. 10-22, 2004.

[20] F. Wang, J. Qiu, J. Yang, B. Dong, X. Li and Y. Li, "Hadoop high availability through metadata replication", CloudDB '09 Proceedings of the 1st international workshop on Cloud data management, pp. 37-44, 2009, ISBN: 978-1-60558-802-5, doi:10.1145/1651263.1651271.

[21] Y. Cao and D. Sun, "Large-scale and big optimization based on hadoop", in Ali Emrouznejad (ed.) "Big data optimization: recent developments and challenges", Springer, ISBN 978-3-319-30265-2, 2016.

[22] C. Ma et al. "Distributed optimization with arbitrary local solvers", ARXIV B.Code: 2015arXiv151204039M, 2015.

[23] M. Jaggi et al., "Communication-efficient distributed dual coordinate ascent" in Advances in Neural Information Processing Systems 27, pp. 3068–3076, 2014.

[24] Y. Zhang, and L. Xiao, "DiSCO: distributed optimization for self-concordant empirical loss" in ICML 2015 - Proceedings of the 32th International Conference on Machine Learning, pp. 362–370, 2015.

[25] C. Jin, C. Vecchiola and R. Buyya, "MRPGA: an extension of mapreduce for parallelizing genetic algorithms", in eScience '08. IEEE 4th Int. Conference on eScience, ISBN 978-1-4244-3380-3, DOI: 10.1109/eScience.2008.78.

[26] F. Teng and D. Tuncay, "Genetic algorithms with mapreduce runtimes", Indiana University Bloomington School of Informatics and Computing Department, unpublished.

[27] J. Lin and C.Dyer, "Data-intensive text processing with mapreduce", Synthesis Lectures on Human Language Tech., Morgan-Claypool, ISBN-13: 978-1608453429, 2010.

[28] M. Zaharia et al., "Resilient distributed datasets: a fault-tolerant abstraction for in-memory computing", 9th USENIX Symposium on Networked Systems Design and Implementation (NSDI 12), pp. 15-28, 2012, San José, California, ISBN 978-931971-92-8.

[29] R. Glowinski, "On alternating direction methods of multipliers: a historical perspective" in W. Fitzgibbon, Y.A. Kuznetsov, P. Neittaanmäki and O. Pironneau (eds.), Vol. 34 Computational Methods in Applied Sciences pp 59-82, "Modeling, Simulation and Optimization for Science and Technology", Springer, 2014.

[30] R.W. Morrison, K.A. De Jong, "A test problem generator for non-stationary environments", in Proceedings of the 1999 Congress on Evolutionary Computation (CEC 1999), vol. 3, pp. 2047–2053, July 1999.

[31] Y. Jin, J. Branke, "Evolutionary optimization in uncertain environments - a survey", IEEE Trans. Evol. Comput. 9(3), pp. 303–317, 2005.

[32] S. Yang, C. Li, "A clustering particle swarm optimizer for locating and tracking multiple optima in dynamic environments", IEEE Trans. Evol. Comput. 14(6), pp. 959–974, 2010.

[33] L.T. Bui, Z. Michalewicz, E. Parkinson, M.B. Abello, "Adaptation in dynamic environments: a case study in mission planning", IEEE Trans. Evol. Comput. 16(2), pp. 190–209, 2012.

[34] C.A.C. Coello, G.B. Lamont, D.A.V. Veldhuizen, "Evolutionary algorithms for solving multi-objective problems" in Genetic and Evolutionary Computation Series, 2nd ed. Springer, New York, 2007.

[35] M. Bhattacharya, R. Islam, and J. Abawajy, "Evolutionary optimization: a big data perspective" Journal of network and computer applications, vol. 59, pp. 416-426, doi: 10.1016/j.jnca.2014.07.032.

[36] E. Cantú-Paz, "A survey of parallel genetic algorithms", Calculateurs Paralleles, Vol. 10, 1998.

[37] J. Cao, H. Cui, H. Shi, and L. Jiao, "Big data: a parallel particle swarm optimization-back-propagation neural network algorithm nased on mapreduce", PLoS ONE 11(6): e0157551. doi:10.1371/journal.pone.0157551, 2016.

[38] J.Q. Feng, W.D. Gu, J.S. Pan, H.J. Zhong, J.D. Huo, "Parallel implementation of BP neural network for traffic prediction on sunway blue light supercomputer" in Applied Mechanics & Materials, Nr. 614, pp. 521–525, 2014.

[39] Y. Kim, K. Shim, M.S. Kim, J.S. Lee, "DBCURE-MR: An efficient density-based clustering algorithm for large data using MapReduce", Information Systems Nr. 42, pp. 15–35, 2013. DOI: 10.1016/j.is.2013.11.002.

[40] J. Qian, P. Lv, X.D. Yue, C.H. Liu, Z.J. Jing, "Hierarchical attribute reduction algorithms for big data using MapReduce", Knowledge-based Systems, Nr. 73, pp. 18–31, 2015, DOI: 10.1016/j.knosys.2014.09.001.

[41] E. Kasturi, S. Prasanna Devi, S. Vinu Kiran, and S. Manivannan, "Airline route profitability analysis and optimization using big data analytics on aviation data sets under heuristic techniques" - Procedia Computer Science 87 (2016), pp. 86–92 -Fourth International Conference on Recent Trends in Computer Science & Engineering.

[42] J. Xu, E. Huang, C. Chen and L. Hay Lee, "Simulation optimization: a review and exploration in the new era of cloud computing and big data", Asia-Pacific Journal of Operational Research Vol. 32, No. 3 (2015) 1550019 (34 pages), World Scientific Publishing Co. & Operational Research Society of Singapore.

[43] J. Xu, S. Zhang, E. Huang, C.H. Chen, L.H. Lee and N. Celik "Efficient multifidelity simulation optimization", Proceedings of 2014 Winter Simulation Conference, 2014.

[44] P. Pääkönen, "Feasibility analysis of AsterixDB and Spark streaming with Cassandra for stream-based processing", Journal of Big Data, 2016.

[45] I. Gorton, A. Basar Bener, A. Mockus, "Software engineering for big data systems", in IEEE Software 33(2), pp. 32-35, March 2016, DOI: 10.1109/MS.2016.47, 2016.

[46] J. Al-Jaroody, B. Hollein, N. Mohamed, "Applying software engineering processes for big data analytics applications development", Conference: 2017 IEEE 7th Annual Computing and Communication Workshop and Conference (CCWC), 2017.

[47] N. H. Madhavji, A.V. Miranskyy, K. Kontogiannis, "Big picture of big data software engineering: with example research challenges", Conference: 2015 IEEE/ACM 1st International Workshop on Big Data Software Engineering (BIGDSE), DOI: 10.1109/BIGDSE.2015.10, 2015.

[48] M. Faisal, T. S. Madeswaran, S. Gupta, "Emphasizing big data engineering and software architecture on cloud computing", (IJCSIT) International Journal of Computer Science and Information Technologies, Vol. 6 (3), pp. 2852-2855, ISSN 0975-9646, 2015.

[49] L. Liu, A. Khodaei, W. Yin, and Z. Han, "A distribute parallel approach for big data scale optimal power flow with security constraints", Conference Paper, Smart Grid Communications (SmartGridComm), 2013 IEEE International Conference.

[50] K. Hall, S. Gilpin, and G. Mann, "MapReduce/Bigtable for distributed optimization" Neural Information Processing Systems Workshop on Leaning on Cores, Clusters, and Clouds, 2010.

[51] N. Marz and J. Warren, "Big data: principles and best practices of scalable realtime data systems", Manning Publications, April 2015, ISBN 9781617290343.

[52] J.Kreps, "Questioning the lambda architecture. The lambda architecture has its merits, but alternatives are worth exploring", O'Reilly Media on line, July 2014, https://www.oreilly.com/ideas/questioning-the-lambda-architecture

[53] J. Durillo, A. Nebro, "jMetal: a java framework for multi-objective optimization", Advances in Engineering Software Nr. 42, pp. 760 – 771, 2011.

[54] C.H. Papadimitriou, "The euclidean travelling salesman problem is NP-Complete", Theoretical Computer Science Nr. 4, pp. 237 – 244, 1977.

[55] K. Shvachko, H. Kuang, S. Radia, R. Chansler, "The Hadoop Distributed File System", MSST '10 Proceedings of the 2010 IEEE 26th Symposium on Mass Storage Systems and Technologies (MSST), pp. 1-10, ISBN: 978-1-4244-7152-2 doi:10.1109/MSST.2010.5496972.

[56] S.Sur, M.J. Koop, D.K. Panda, "High-performance and scalable MPI over InfiniBand with reduced memory usage: an in-depth performance analysis", SC '06 Proceedings of the 2006 ACM/IEEE conference on Supercomputing Article No. 105, ISBN: 0-7695-2700-0, 2006.

[57] T. Sterling, D.J. Becker, D. Savarese, J.E. Dorband, U. A. Ranawake and C. V. Packer, "Beowulf: a parallel workstation for scientific computation", in Proceedings of the 24th International Conference on Parallel Processing, pp. 11-14, 1995.

[58] D. Whiteson, "Higgs data set", Machine Learning Repository, Center for Machine Learning and Intelligent Systems, UCI, on line at March 2017, published online at https://archive.ics.uci.edu/ml/datasets/HIGGS 2014.